\newcommand{\omt}[1]{}
\documentclass[11pt]{article}
\usepackage{colacl, epsfig}

\title{Supervised Grammar Induction using Training Data with Limited
Constituent Information
\thanks{This material is based upon work supported by the National
Science Foundation under Grant No. IRI 9712068.  We thank Stuart
Shieber for his guidance, and Lillian Lee, Ric Crabbe, and the three
anonymous reviewers for their comments on the paper.}}
\author{Rebecca Hwa                            \\
  Division of Engineering and Applied Sciences \\
  Harvard  University                          \\ 
  Cambridge, MA 02138 USA                      \\
  rebecca@eecs.harvard.edu
}
	
\begin{document}
\maketitle
\begin{abstract}

Corpus-based grammar induction generally relies on hand-parsed
training data to learn the structure of the language.  Unfortunately,
the cost of building large annotated corpora is prohibitively
expensive.  This work aims to improve the induction strategy when
there are few labels in the training data.  We show that the most
informative linguistic constituents are the higher nodes in the parse
trees, typically denoting complex noun phrases and sentential clauses.
They account for only 20\% of all constituents.  For inducing grammars
from sparsely labeled training data (e.g., only higher-level
constituent labels), we propose an {\it adaptation} strategy, which
produces grammars that parse almost as well as grammars induced from
fully labeled corpora.  Our results suggest that for a partial parser
to replace human annotators, it must be able to automatically extract
higher-level constituents rather than base noun phrases.
\end{abstract}

\bibliographystyle{acl}

\section{Introduction}

The availability of large hand-parsed corpora such as the Penn
Treebank Project has made high-quality statistical parsers possible.
However, the parsers risk becoming too tailored to these labeled
training data that they cannot reliably process sentences from an
arbitrary domain.  Thus, while a parser trained on the Wall Street
Journal corpus can fairly accurately parse a new Wall Street Journal
article, it may not perform as well on a New Yorker article.  To parse
sentences from a new domain, one would normally {\it directly induce}
a new grammar from that domain, in which the training process would
require hand-parsed sentences from the new domain.  Because parsing a
large corpus by hand is a labor-intensive task, it would be beneficial
to minimize the number of labels needed to induce the new grammar.

We propose to {\it adapt} a grammar already trained on an old domain
to the new domain.  Adaptation can exploit the structural similarity
between the two domains so that fewer labeled data might be needed to
update the grammar to reflect the structure of the new domain.  This
paper presents a quantitative study comparing direct induction and
adaptation under different training conditions.  Our goal is to
understand the effect of the amounts and types of labeled data on the
training process for both induction strategies.  For example, how much
training data need to be hand-labeled?  Must the parse trees for each
sentence be fully specified?  Are some linguistic constituents in the
parse more informative than others?  

To answer these questions, we have performed experiments that compare
the parsing qualities of grammars induced under different training
conditions using both adaptation and direct induction.  We vary the
number of labeled brackets and the linguistic classes of the labeled
brackets.  The study is conducted on both a simple Air Travel
Information System (ATIS) corpus \cite{Hemphill:90a} and the more
complex Wall Street Journal (WSJ) corpus
\cite{Marcus:93a}.

Our results show that the training examples do not need to be fully
parsed for either strategy, but adaptation produces better grammars
than direct induction under the conditions of minimally labeled
training data.  For instance, the most informative brackets, which
label constituents higher up in the parse trees, typically identifying
complex noun phrases and sentential clauses, account for only 17\% of
all constituents in ATIS and 21\% in WSJ.  Trained on this type of
label, the adapted grammars parse better than the directly induced
grammars and almost as well as those trained on fully labeled data.
Training on ATIS sentences labeled with higher-level constituent
brackets, a directly induced grammar parses test sentences with 66\%
accuracy, whereas an adapted grammar parses with 91\% accuracy, which
is only 2\% lower than the score of a grammar induced from fully
labeled training data.  Training on WSJ sentences labeled with
higher-level constituent brackets, a directly induced grammar parses
with 70\% accuracy, whereas an adapted grammar parses with 72\%
accuracy, which is 6\% lower than the score of a grammar induced from
fully labeled training data.

That the most informative brackets are higher-level constituents and
make up only one-fifth of all the labels in the corpus has two
implications.  First, it shows that there is potential reduction of
labor for the human annotators.  Although the annotator still must
process an entire sentence mentally, the task of identifying
higher-level structures such as sentential clauses and complex nouns
should be less tedious than to fully specify the complete parse tree
for each sentence.  Second, one might speculate the possibilities of
replacing human supervision altogether with a partial parser that
locates constituent chunks within a sentence.  However, as our results
indicate that the most informative constituents are higher-level
phrases, the parser would have to identify sentential clauses and
complex noun phrases rather than low-level base noun phrases.

\section{Related Work on Grammar Induction}

Grammar induction is the process of inferring the structure of a
language by learning from example sentences drawn from the language.
The degree of difficulty in this task depends on three factors.
First, it depends on the amount of supervision provided.
\newcite{Charniak:96b}, for instance, has shown that a grammar can be
easily constructed when the examples are fully labeled parse trees.
On the other hand, if the examples consist of raw sentences with no
extra structural information, grammar induction is very difficult,
even theoretically impossible \cite{Gold:67}.  One could take a greedy
approach such as the well-known Inside-Outside re-estimation algorithm
\cite{Baker:79b}, which induces locally optimal grammars by
iteratively improving the parameters of the grammar so that the
entropy of the training data is minimized.  In practice, however, when
trained on unmarked data, the algorithm tends to converge on poor
grammar models.  For even a moderately complex domain such as the ATIS
corpus, a grammar trained on data with constituent bracketing
information produces much better parses than one trained on completely
unmarked raw data \cite{Pereira:92a}.  Part of our work explores the
in-between case, when only some constituent labels are available.
Section \ref{brktypes} defines the different types of annotation we
examine.

Second, as supervision decreases, the learning process relies more on
search.  The success of the induction depends on the initial
parameters of the grammar because a local search strategy may
converge to a local minimum.  For finding a good initial
parameter set, \newcite{Lari:90a} suggested first estimating the
probabilities with a set of regular grammar rules.  Their experiments,
however, indicated that the main benefit from this type of
pretraining is one of run-time efficiency; the improvement in the
quality of the induced grammar was minimal.  \newcite{Briscoe:92}
argued that one should first hand-design the grammar to encode some
linguistic notions and then use the re-estimation procedure to
fine-tune the parameters, substituting the cost of hand-labeled
training data with that of hand-coded grammar.  Our idea of grammar
adaptation can be seen as a form of initialization.  It attempts to
seed the grammar in a favorable search space by first training it with
data from an existing corpus.  Section \ref{gtypes} discusses the
induction strategies in more detail.

\begin{table*}[t]
\begin{tabular}
{|l|p{4.3in}|c|c|} \hline
\bf Categories & \bf Labeled Sentence                             & \bf ATIS & \bf WSJ \\ \hline \hline
HighP  & (I want (to take (the flight with at most one stop)))           & 17\% & 21\% \\ \hline
BaseNP & (I) want to take (the flight) with (at most one stop)             & 27\% & 29\% \\ \hline
BaseP  & (I) want to take (the flight) with (at most one) stop             & 32\% & 30\% \\ \hline
AllNP  & (I) want to take ((the flight) with (at most one stop))           & 37\% & 43\% \\ \hline
NotBaseP & (I (want (to (take (the flight (with (at most one stop))))))) & 68\% & 70\% \\ \hline 
\end{tabular}
\caption{The second column shows how the example sentence  (($I$)
($want$ ($to$ ($take$ (($the$ $flight$) ($with$ (($at$ $most$ $one$)
$stop$))))))) is labeled under each category.  The third and fourth
columns list the percentage break-down of brackets in each category
for ATIS and WSJ respectively.}
\label{brktype:tab}
\end{table*}

A third factor that affects the learning process is the complexity of
the data.  In their study of parsing the WSJ,
\newcite{Schabes:93a} have shown that a grammar trained on the
Inside-Outside re-estimation algorithm can perform quite well on short
simple sentences but falters as the sentence length increases.  To
take this factor into account, we perform our experiments on both a
simple domain (ATIS) and a complex one (WSJ).  In Section \ref{exps},
we describe the experiments and report the results.

\section{Training Data Annotation}
\label{brktypes}

The training sets are annotated in multiple ways, falling into two
categories.  First, we construct training sets annotated with random
subsets of constituents consisting 0\%, 25\%, 50\%, 75\% and 100\% of
the brackets in the fully annotated corpus.  Second, we construct sets
training in which only a certain type of constituent is annotated.  We
study five linguistic categories.  Table
\ref{brktype:tab} summarizes the annotation differences between the
five classes and lists the percentage of brackets in each class with
respect to the total number of constituents\footnote{For computing the
percentage of brackets, the outer-most bracket around the entire
sentence and the brackets around singleton phrases (e.g., the pronoun
``I'' as a BaseNP) are excluded because they do not contribute to the
pruning of parses.}  for ATIS and WSJ.  In an {\bf AllNP} training
set, all and only the noun phrases in the sentences are labeled.  For
the {\bf BaseNP} class, we label only simple noun phrases that contain
no embedded noun phrases. Similarly for a {\bf BaseP} set, all simple
phrases made up of only lexical items are labeled.  Although there is
a high intersection between the set of BaseP labels and the set of
BaseNP labels, the two classes are not identical.  A BaseNP may
contain a BaseP.  For the example in Table
\ref{brktype:tab}, the phrase ``at most one stop'' is a BaseNP that
contains a quantifier BaseP ``at most one.''  {\bf NotBaseP} is the
complement of BaseP. The majority of the constituents in a sentence
belongs to this category, in which at least one of the constituent's
sub-constituents is not a simple lexical item.  Finally, in a {\bf
HighP} set, we label only complex phrases that decompose into
sub-phrases that may be either another HighP or a BaseP.  That is, a
HighP constituent does not directly subsume any lexical word.  A
typical HighP is a sentential clause or a complex noun phrase.  The
example sentence in Table
\ref{brktype:tab} contains 3 HighP constituents: a complex noun phrase
made up of a BaseNP and a prepositional phrase; a sentential
clause with an omitted subject NP; and the full sentence.

\section{Induction Strategies}
\label{gtypes}

To induce a grammar from the sparsely bracketed training data
previously described, we use a variant of the Inside-Outside
re-estimation algorithm proposed by \newcite{Pereira:92a}.  The
inferred grammars are represented in the Probabilistic Lexicalized
Tree Insertion Grammar (PLTIG) formalism \cite{Schabes:93b,Hwa:98b}, which is
lexicalized and context-free equivalent.  We favor the PLTIG
representation for two reasons. First, it is amenable to the
Inside-Outside re-estimation algorithm (the equations calculating the
inside and outside probabilities for PLTIGs can be found in
\newcite{Hwa:98a}).  Second, its lexicalized representation makes the
training process more efficient than a traditional PCFG while
maintaining comparable parsing qualities.

Two training strategies are considered: direct induction, in which a
grammar is induced from scratch, learning from only the sparsely
labeled training data; and adaptation, a two-stage learning process
that first uses direct induction to train the grammar on an existing
fully labeled corpus before retraining it on the new corpus.  During
the retraining phase, the probabilities of the grammars are
re-estimated based on the new training data.  We expect the adaptive
method to induce better grammars than direct induction when the new
corpus is only partially annotated because the adapted grammars have
collected better statistics from the fully labeled data of another
corpus.

\section{Experiments and Results}
\label{exps}
We perform two experiments.  The first uses ATIS as the corpus from
which the different types of partially labeled training sets are
generated.  Both induction strategies train from these data, but the
adaptive strategy pretrains its grammars with fully labeled data
drawn from the WSJ corpus.  The trained grammars are scored on their
parsing abilities on unseen ATIS test sets.  We use the non-crossing
bracket measurement as the parsing metric.  This experiment will show
whether annotations of a particular linguistic category may be more
useful for training grammars than others.  It will also indicate the
comparative merits of the two induction strategies trained on data
annotated with these linguistic categories.  However, pretraining on
the much more complex WSJ corpus may be too much of an advantage for
the adaptive strategy.  Therefore, we reverse the roles of the corpus
in the second experiment.  The partially labeled data are from the WSJ
corpus, and the adaptive strategy is pretrained on fully labeled ATIS
data.  In both cases, part-of-speech(POS) tags are used as the lexical
items of the sentences.  Backing off to POS tags is necessary because
the tags provide a considerable intersection in the vocabulary
sets of the two corpora.

\subsection{Experiment 1: Learning ATIS}
\label{w2a}

\begin{figure*}[t]
\begin{center}
\begin{tabular}{cc}
	\mbox{\psfig{figure=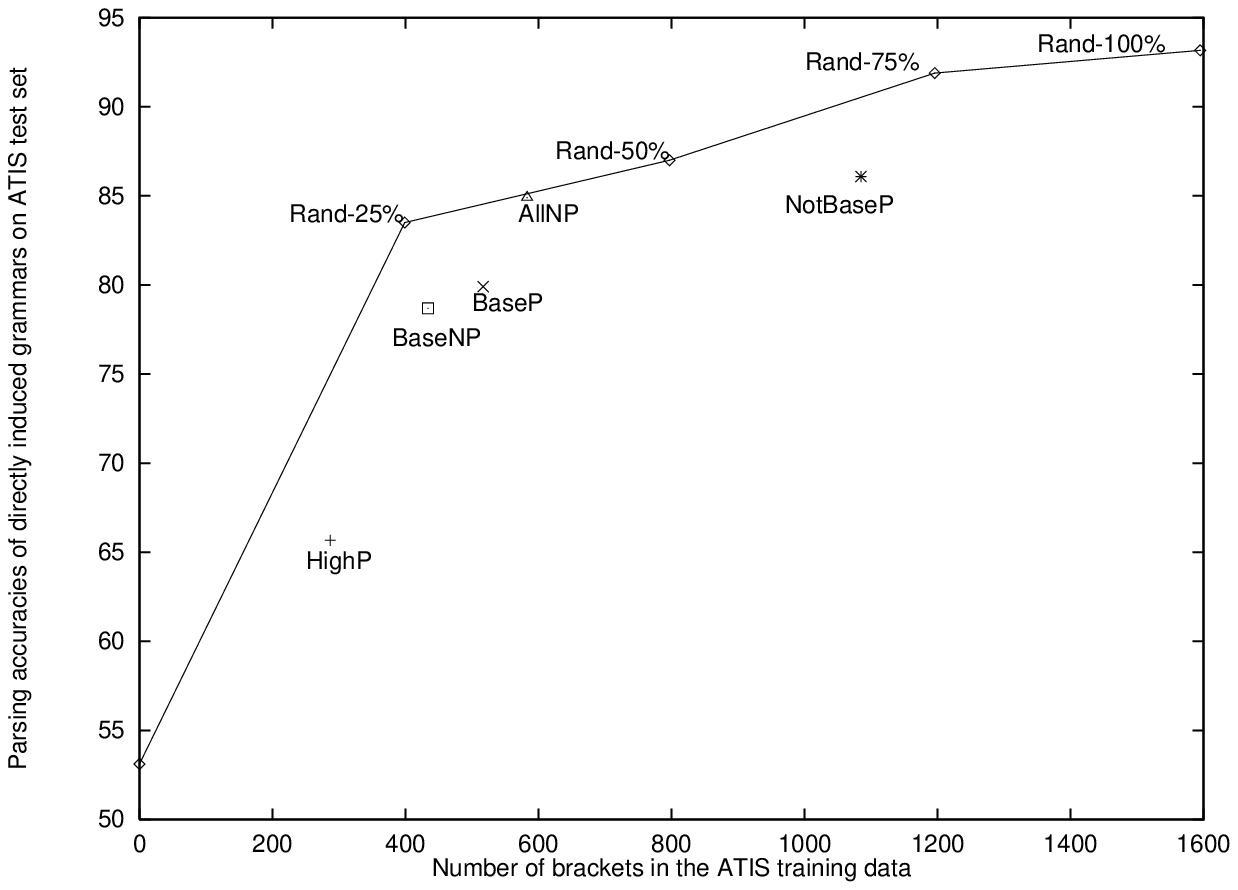,width=3in}}
&   \mbox{\psfig{figure=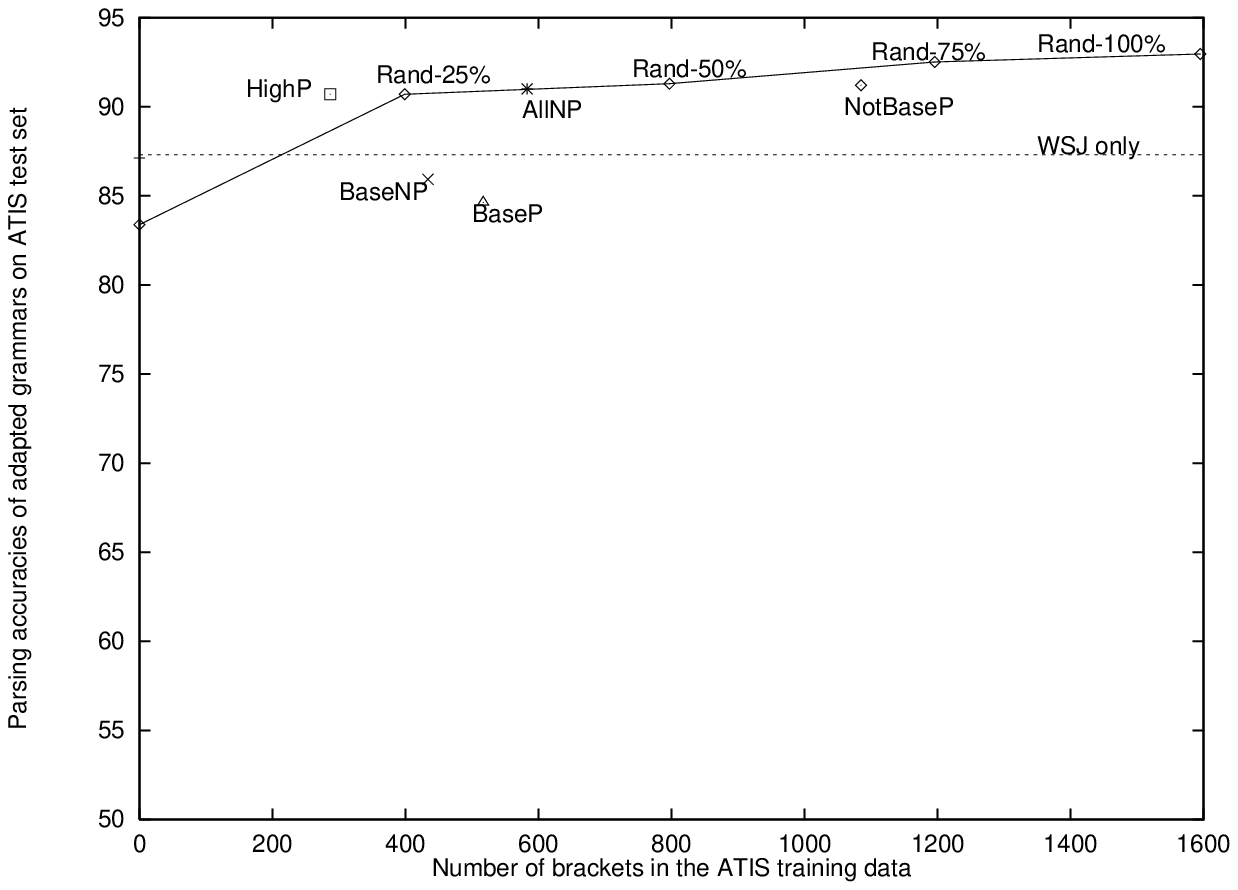,width=3in}} \\
(a) & (b)
\end{tabular}
\end{center}
\caption{Parsing accuracies of (a) directly induced grammars and (b)
adapted grammars as a function of the number of brackets present in
the training corpus.  There are 1595 brackets in the training corpus
all together.}
\label{fig:w2a.out}
\end{figure*}

The easier learning task is to induce grammars to parse ATIS
sentences.  The ATIS corpus consists of 577 short sentences with
simple structures, and the vocabulary set is made up of 32 POS tags, a
subset of the 47 tags used for the WSJ.  Due to the limited size of
this corpus, ten sets of randomly partitioned train-test-held-out
triples are generated to ensure the statistical significance of our
results.  We use 80 sentences for testing, 90 sentences for held-out
data, and the rest for training.  Before proceeding with the main
discussion on training from the ATIS, we briefly describe the
pretraining stage of the adaptive strategy.

\subsubsection{Pretraining with WSJ}

The idea behind the adaptive method is simply to make use of any
existing labeled data.  We hope that pretraining the grammars on
these data might place them in a better position to learn from the
new, sparsely labeled data.  In the pretraining stage for this
experiment, a grammar is directly induced from 3600 fully labeled WSJ
sentences.  Without any further training on ATIS data, this grammar
achieves a parsing score of 87.3\% on ATIS test sentences.  The
relatively high parsing score suggests that pretraining with WSJ has 
successfully placed the grammar in a good position to begin training
with the ATIS data.  

\subsubsection{Partially Supervised Training on ATIS}

We now return to the main focus of this experiment: learning from
sparsely annotated ATIS training data.  To verify whether some
constituent classes are more informative than others, we could compare
the parsing scores of the grammars trained using different constituent
class labels.  But this evaluation method does not take into account
that the distribution of the constituent classes is not uniform.  To
normalize for this inequity, we compare the parsing scores to a
baseline that characterizes the relationship between the performance
of the trained grammar and the number of bracketed constituents in the
training data.  To generate the baseline, we create training data in
which 0\%, 25\%, 50\%, 75\%, and 100\% of the constituent brackets are
randomly chosen to be included.  One class of linguistic labels is
better than another if its resulting parsing improvement over the
baseline is higher than that of the other.

The test results of the grammars induced from these different training
data are summarized in Figure \ref{fig:w2a.out}.  Graph (a) plots the
outcome of using the direct induction strategy, and graph (b) plots
the outcome of the adaptive strategy.  In each graph, the baseline of
random constituent brackets is shown as a solid line.  Scores of
grammars trained from constituent type specific data sets are plotted
as labeled dots.  The dotted horizontal line in graph (b) indicates
the ATIS parsing score of the grammar trained on WSJ alone.

Comparing the five constituent types, we see that the HighP class is
the most informative for the adaptive strategy, resulting in a grammar
that scored better than the baseline.  The grammars trained on the
AllNP annotation performed as well as the baseline for both
strategies.  Grammars trained under all the other training conditions
scored below the baseline.  Our results suggest that while an ideal
training condition would include annotations of both higher-level
phrases and simple phrases, complex clauses are more informative.
This interpretation explains the large gap between the parsing scores
of the directly induced grammar and the adapted grammar trained on
the same HighP data.  The directly induced grammar performed poorly
because it has never seen a labeled example of simple phrases.  In
contrast, the adapted grammar was already exposed to labeled WSJ
simple phrases, so that it successfully adapted to the new corpus from
annotated examples of higher-level phrases.  On the other hand,
training the adapted grammar on annotated ATIS simple phrases is not
successful even though it has seen examples of WSJ higher-level
phrases.  This also explains why grammars trained on the conglomerate
class NotBaseP performed on the same level as those trained on the
AllNP class.  Although the NotBaseP set contains the most brackets,
most of the brackets are irrelevant to the training process, as they
are neither higher-level phrases nor simple phrases.

Our experiment also indicates that induction strategies exhibit
different learning characteristics under partially supervised training
conditions.  A side by side comparison of Figure \ref{fig:w2a.out}
(a) and (b) shows that the adapted grammars perform significantly
better than the directly induced grammars as the level of supervision
decreases.  This supports our hypothesis that pretraining on a
different corpus can place the grammar in a good initial search space
for learning the new domain.  Unfortunately, a good initial state does
not obviate the need for supervised training.  We see from Figure
\ref{fig:w2a.out}(b) that retraining with unlabeled ATIS sentences
actually lowers the grammar's parsing accuracy.

\subsection{Experiment 2: Learning WSJ}
\label{a2w}

In the previous section, we have seen that annotations of complex
clauses are the most helpful for inducing ATIS-style grammars.  One
of the goals of this experiment is to verify whether the result also
holds for the WSJ corpus, which is structurally very different from
ATIS.  The WSJ corpus uses 47 POS tags, and its sentences are longer
and have more embedded clauses.  

As in the previous experiment, we construct training sets with
annotations of different constituent types and of different numbers of
randomly chosen labels.  Each training set consists of 3600 sentences,
and 1780 sentences are used as held-out data.  The trained grammars
are tested on a set of 2245 sentences.

\begin{figure*}[t]
\begin{center}
\begin{tabular}{cc}
	\mbox{\psfig{figure=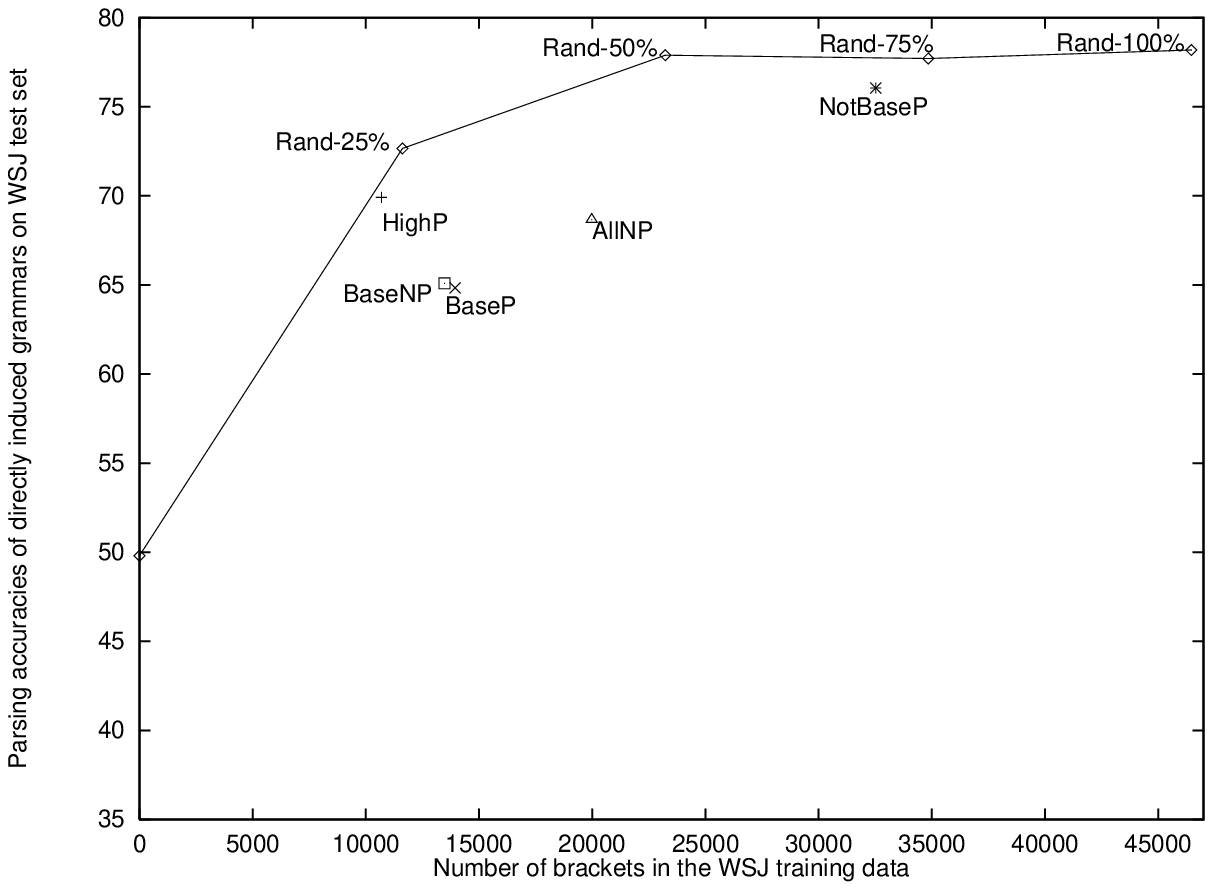,width=3in}}
&   \mbox{\psfig{figure=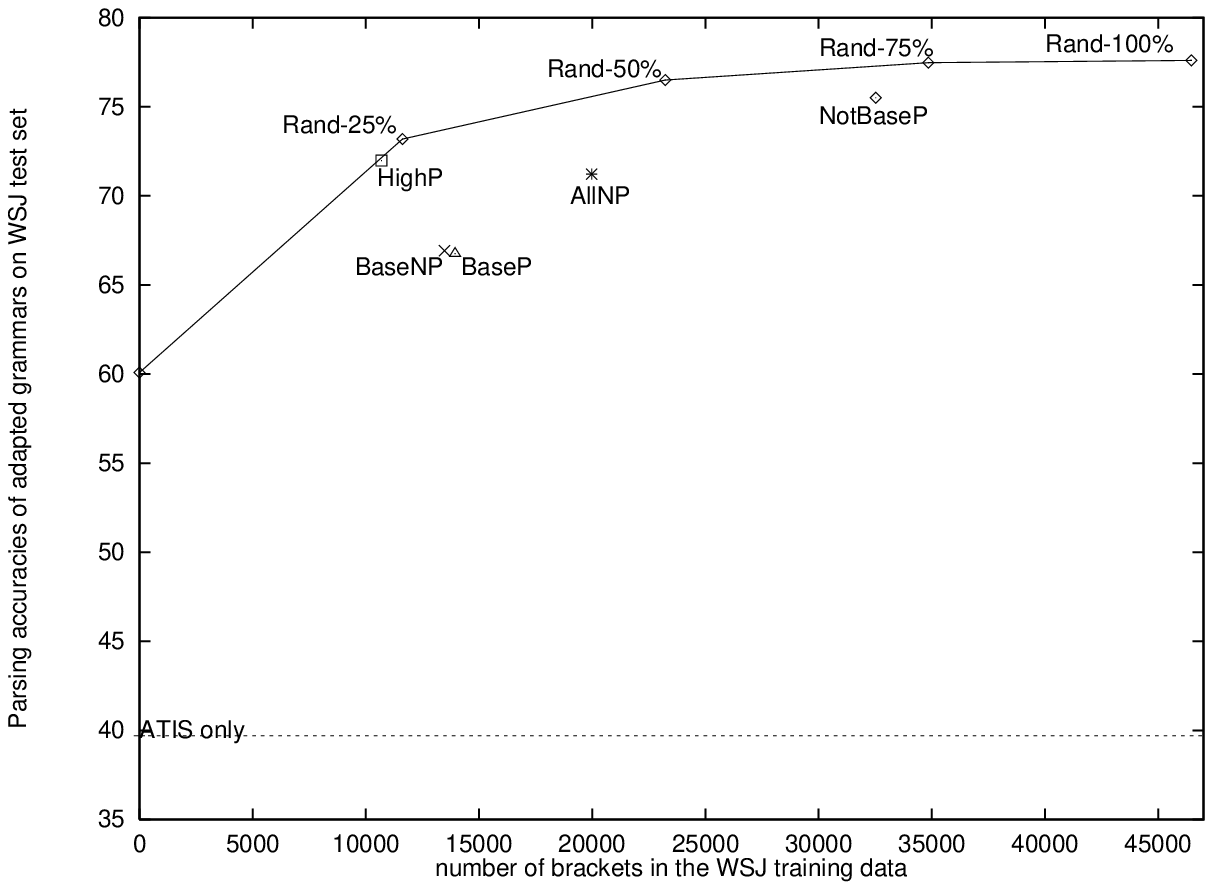,width=3in}} \\
(a) & (b)
\end{tabular}
\end{center}
\caption{Parsing accuracies of (a) directly induced grammars and (b)
adapted grammars as a function of the number of brackets present in
the training corpus.  There is a total of 46463 brackets in the
training corpus.}  
\label{fig:a2w.out}
\end{figure*}

Figure \ref{fig:a2w.out} (a) and (b) summarize the outcomes of this
experiment.  Many results of this section are similar to the ATIS
experiment.  Higher-level phrases still provide the most information;
the grammars trained on the HighP labels are the only ones that scored
as well as the baseline.  Labels of simple phrases still seem the
least informative; scores of grammars trained on BaseP and BaseNP
remained far below the baseline.  Different from the previous
experiment, however, the AllNP training sets do not seem to provide as
much information for this learning task.  This may be due to the
increase in the sentence complexity of the WSJ, which further
de-emphasized the role of the simple phrases.  Thus, grammars trained
on AllNP labels have comparable parsing scores to those trained on
HighP labels.  Also, we do not see as big a gap between the scores of
the two induction strategies in the HighP case because the adapted
grammar's advantage of having seen annotated ATIS base nouns is
reduced.  Nonetheless, the adapted grammars still perform 2\% better
than the directly induced grammars, and this improvement is
statistically significant.\footnote{A pair-wise t-test comparing the
parsing scores of the ten test sets for the two strategies shows 99\%
confidence in the difference.}

Furthermore, grammars trained on NotBaseP
do not fall as far below the baseline and have higher parsing scores
than those trained on HighP and AllNP.  This suggests that for more
complex domains, other linguistic constituents such as verb phrases\footnote{We have not experimented with training sets containing only verb
phrases labels (i.e., setting a pair of bracket around the head verb
and its modifiers).  They are a subset of the NotBaseP class.} 
become more informative.

A second goal of this experiment is to test the adaptive strategy
under more stringent conditions.  In the previous experiment, a
WSJ-style grammar was retrained for the simpler ATIS corpus.  Now,
we reverse the roles of the corpora to see whether the adaptive
strategy still offers any advantage over direct induction.  

In the adaptive method's pretraining stage, a grammar is induced from
400 fully labeled ATIS sentences.  Testing this ATIS-style grammar on
the WSJ test set without further training renders a parsing accuracy
of 40\%.  The low score suggests that fully labeled ATIS data does not
teach the grammar as much about the structure of WSJ.  Nonetheless,
the adaptive strategy proves to be beneficial for learning WSJ from
sparsely labeled training sets.  The adapted grammars out-perform the
directly induced grammars when more than 50\% of the brackets are
missing from the training data.  The most significant difference is
when the training data contains no label information at all.  The
adapted grammar parses with 60.1\% accuracy whereas the directly
induced grammar parses with 49.8\% accuracy.  

\section{Conclusion and Future Work}

In this study, we have shown that the structure of a grammar can be
reliably learned without having fully specified constituent
information in the training sentences and that the most informative
constituents of a sentence are higher-level phrases, which make up
only a small percentage of the total number of constituents.  
Moreover, we observe that grammar adaptation works particularly well
with this type of sparse but informative training data.  An adapted
grammar consistently outperforms a directly induced grammar even
when adapting from a simpler corpus to a more complex one.

These results point us to three future directions.  First, that the
labels for some constituents are more informative than others implies
that sentences containing more of these informative constituents make
better training examples.  It may be beneficial to estimate the
informational content of potential training (unmarked) sentences.  The
training set should only include sentences that are predicted to have
high information values.  Filtering out unhelpful sentences from the
training set reduces unnecessary work for the human annotators.
Second, although our experiments show that a sparsely labeled training
set is more of an obstacle for the direct induction approach than for
the grammar adaptation approach, the direct induction strategy might
also benefit from a two stage learning process similar to that used
for grammar adaptation.  Instead of training on a different corpus in
each stage, the grammar can be trained on a small but fully labeled
portion of the corpus in its first stage and the sparsely labeled
portion in the second stage.  Finally, higher-level constituents have
proved to be the most informative linguistic units.  To relieve humans
from labeling any training data, we should consider using partial
parsers that can automatically detect complex nouns and sentential
clauses.

\bibliography{partial}

\begin{thebibliography}{}

\bibitem[\protect\citename{Baker}1979]{Baker:79b}
J.K. Baker.
\newblock 1979.
\newblock Trainable grammars for speech recognition.
\newblock In {\em Proceedings of the Spring Conference of the Acoustical
  Society of America}, pages 547--550, Boston, MA, June.

\bibitem[\protect\citename{Briscoe and Waegner}1992]{Briscoe:92}
E.J. Briscoe and N.~Waegner.
\newblock 1992.
\newblock Robust stochastic parsing using the inside-outside algorithm.
\newblock In {\em Proceedings of the AAAI Workshop on Probabilistically-Based
  NLP Techniques}, pages 39--53.

\bibitem[\protect\citename{Charniak}1996]{Charniak:96b}
E.~Charniak.
\newblock 1996.
\newblock Tree--bank grammars.
\newblock In {\em Proceedings of the Thirteenth National Conference on
  Artificial Intelligence}, pages 1031--1036.

\bibitem[\protect\citename{Gold}1967]{Gold:67}
E.~Mark Gold.
\newblock 1967.
\newblock Language identification in the limit.
\newblock {\em Information Control}, 10(5):447--474.

\bibitem[\protect\citename{Hemphill \bgroup et al.\egroup }1990]{Hemphill:90a}
C.T. Hemphill, J.J. Godfrey, and G.R. Doddington.
\newblock 1990.
\newblock The {ATIS} spoken language systems pilot corpus.
\newblock In {\em DARPA Speech and Natural Language Workshop}, Hidden Valley,
  Pennsylvania, June. Morgan Kaufmann.

\bibitem[\protect\citename{Hwa}1998a]{Hwa:98b}
R.~Hwa.
\newblock 1998a.
\newblock An empirical evaluation of probabilistic lexicalized tree insertion
  grammars.
\newblock In {\em Proceedings of COLING-ACL}, volume~1, pages 557--563.

\bibitem[\protect\citename{Hwa}1998b]{Hwa:98a}
R.~Hwa.
\newblock 1998b.
\newblock An empirical evaluation of probabilistic lexicalized tree insertion
  grammars.
\newblock Technical Report 06-98, Harvard University.
\newblock Available as cmp-lg/9808001.

\bibitem[\protect\citename{Lari and Young}1990]{Lari:90a}
K.~Lari and S.J. Young.
\newblock 1990.
\newblock The estimation of stochastic context-free grammars using the
  inside-outside algorithm.
\newblock {\em Computer Speech and Language}, 4:35--56.

\bibitem[\protect\citename{Marcus \bgroup et al.\egroup }1993]{Marcus:93a}
M.~Marcus, B.~Santorini, and M.~Marcinkiewicz.
\newblock 1993.
\newblock Building a large annontated corpus of english: the penn treebank.
\newblock {\em Computational Linguistics}, 19(2):313--330.

\bibitem[\protect\citename{Pereira and Schabes}1992]{Pereira:92a}
F.~Pereira and Y.~Schabes.
\newblock 1992.
\newblock {I}nside-{O}utside reestimation from partially bracketed corpora.
\newblock In {\em Proceedings of the 30th Annual Meeting of the ACL}, pages
  128--135, Newark, Delaware.

\bibitem[\protect\citename{Schabes and Waters}1993]{Schabes:93b}
Y.~Schabes and R.~Waters.
\newblock 1993.
\newblock Stochastic lexicalized context-free grammar.
\newblock In {\em Proceedings of the Third International Workshop on Parsing
  Technologies}, pages 257--266.

\bibitem[\protect\citename{Schabes \bgroup et al.\egroup }1993]{Schabes:93a}
Y.~Schabes, M.~Roth, and R.~Osborne.
\newblock 1993.
\newblock Parsing the {W}all {S}treet {J}ournal with the {I}nside-{O}utside
  algorithm.
\newblock In {\em Proceedings of the Sixth Conference of the European Chapter
  of the ACL}, pages 341--347.

\end{thebibliography}
\end{document}